\documentclass[11pt,a4paper]{article}
\usepackage[hyperref]{eacl2021}
\usepackage{times}
\usepackage{latexsym}

\usepackage{courier}
\usepackage{caption}
\usepackage{natbib}
\usepackage{graphicx}
\usepackage{latexsym}
\usepackage{amsmath}
\usepackage{url}
\usepackage{mathtools}
\usepackage{mathdots}
\usepackage{color}
\usepackage{siunitx}
\usepackage{array}
\usepackage{multirow}
\usepackage{amssymb}
\usepackage{gensymb}
\usepackage{tabularx}
\usepackage{booktabs}
\usepackage{cleveref}
\usepackage{xcolor}
\usepackage{makecell}
\usepackage{multirow}
\usepackage[normalem]{ulem}  
\useunder{\uline}{\ul}{}

\usepackage{microtype}

\aclfinalcopy 

\title{Zero-shot Generalization in Dialog State Tracking through Generative Question Answering}

\author{
    Shuyang Li,\textsuperscript{\rm 1, 2} 
    Jin Cao,\textsuperscript{\rm 1} 
    Mukund Sridhar,\textsuperscript{\rm 1} 
    Henghui Zhu,\textsuperscript{\rm 3} 
    Shang-Wen Li,\textsuperscript{\rm 3}
    \\
    \textbf{Wael Hamza},\textsuperscript{\rm 1}
    \textbf{Julian McAuley}\textsuperscript{\rm 2}
    \\
    \textsuperscript{\rm 1}Amazon Alexa AI, \textsuperscript{\rm 2}UC San Diego, \textsuperscript{\rm 3}AWS AI\\
    \texttt{\{shl008, jmcauley\}@eng.ucsd.edu}\\
    \texttt{\{jincao, harakere, henghui, shangwel, waelhamz\}@amazon.com}
}

\date{}

\begin{document}
\maketitle
\begin{abstract}
Dialog State Tracking (DST), an integral part of modern dialog systems, aims to track user preferences and constraints (slots) in task-oriented dialogs.
In real-world settings with constantly changing services, DST systems must generalize to new domains and unseen slot types.
Existing methods for DST do not generalize well to new slot names and many require known ontologies of slot types and values for inference.
We introduce a novel ontology-free framework that supports natural language queries for unseen constraints and slots in multi-domain task-oriented dialogs.
Our approach is based on generative question-answering using a conditional language model pre-trained on substantive English sentences.
Our model improves joint goal accuracy in zero-shot domain adaptation settings by up to 9\% (absolute) over the previous state-of-the-art on the MultiWOZ 2.1 dataset.
\end{abstract}

\section{Introduction}
Dialog agents are gaining increasing prominence in daily life.
These systems aim to assist users via natural language conversations, taking the form of digital assistants who help accomplish everyday tasks by interfacing with connected devices and services.
A key component to understanding and enabling these task-oriented dialogs is Dialog State Tracking (DST): extracting user intent and goals from conversations via filling in belief slots \citep{DBLP:conf/eacl/LemonGHS06,DBLP:conf/sigdial/WangL13}.
Assistive and recommendation use-cases for dialog agents in production settings are particularly challenging due to constantly changing services and applications with which they interface.

Traditional DST systems have achieved high accuracy when presented with a known ontology of slot types and valid values \citep{DBLP:conf/aaai/0002LWZT020}.
In a real-world setting, however, a DST model must generalize to new slot \emph{values} (e.g.~new entities that are not present at training time) and new slot \emph{types} (e.g.~requirements regarding a new application).
Recent work has sought to address these issues by posing DST as a reading comprehension or question answering (QA) task \citep{DBLP:conf/sigdial/GaoSACH19}---such models predict each slot value independently at any given turn and can theoretically be queried for new slots at inference time.

\begin{figure}[t!]
\centering
\includegraphics[width=0.99\columnwidth]{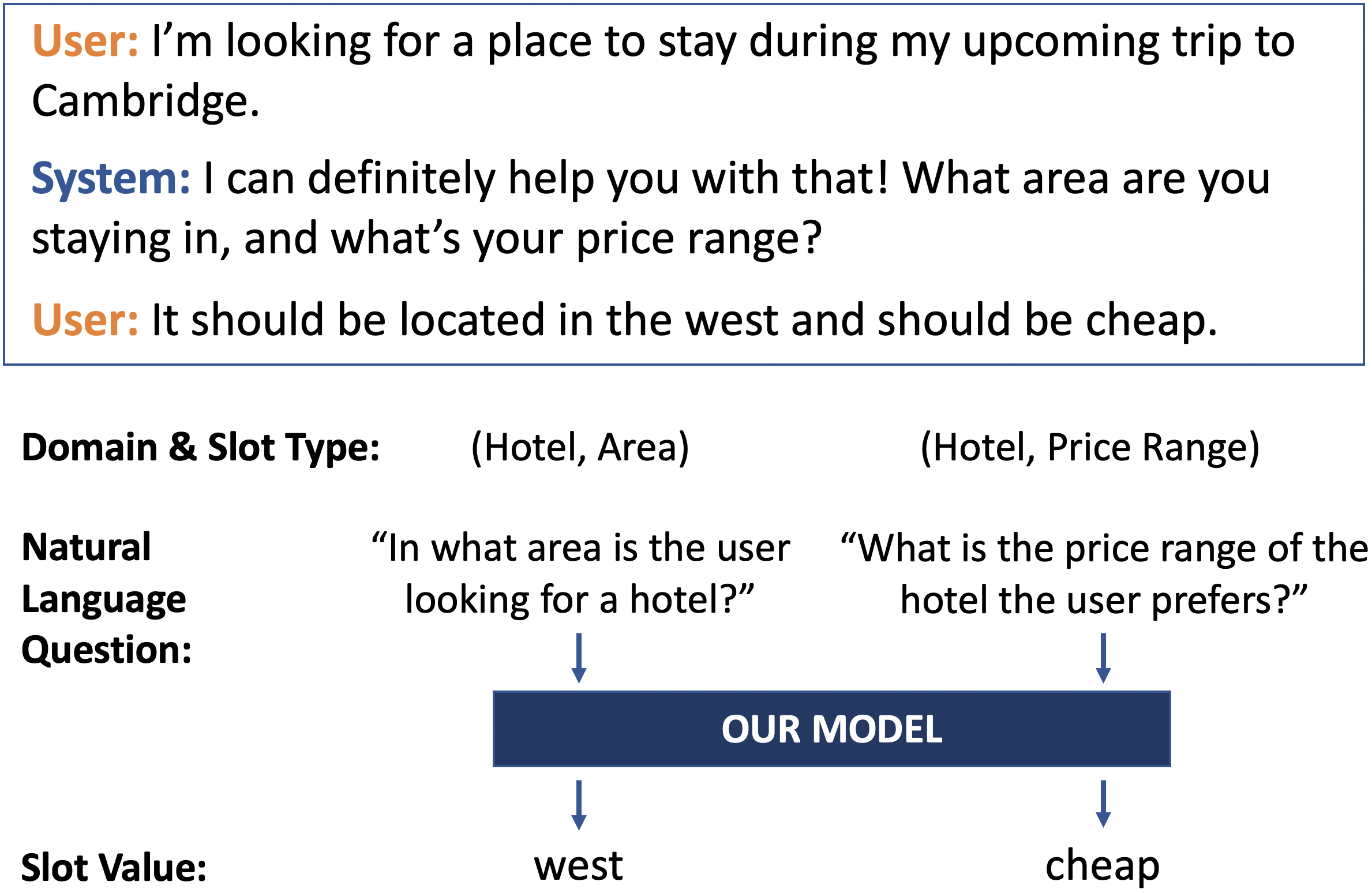}
\caption{Based on a dialog history, a natural language questions are provided to our model to query a user's requirements and preferences (dialog state).
}
\label{fig:problem_setup}
\end{figure}

Some approaches toward DST as QA learn embedding vectors for each slot and/or domain word \citep{DBLP:conf/acl/WuMHXSF19}, but this is not robust to unseen slots whose specific names (e.g.~`Internet Access') may be totally unlike those in the training set.
\citet{DBLP:journals/corr/abs-2004-05827} attempt to remedy this by posing a natural language question for each slot, but their hybrid span-extraction and classification-based system nonetheless requires access to the full ontology for unknown domains.
We present an ontology-free model using natural language questions to represent slots that builds on conditional language modeling techniques---taking advantage of the rise of powerful generative language models \citep{radford2019better}---to tackle DST as a \emph{generative} QA task.
Our model can generalize to unseen domains, slot types, and values, and allows developers to query for arbitrary user requirements via simple questions.
To summarize our main contributions:
\begin{itemize}
    \item We propose an ontology-free conditional language modeling framework for dialog state tracking via generative question answering, achieving state-of-the-art performance in zero-shot domain adaptation settings for DST on MultiWOZ 2.1 \cite{DBLP:conf/lrec/EricGPSAGKGKH20} across all domains with average per-domain gains of 5.9\% joint accuracy over previous best methods;
    \item We demonstrate performance competitive with state-of-the-art methods in a fully supervised setting;
    \item We show that our approach can be easily adapted to predict slot carry-over and transfer knowledge from a larger, more diverse dataset \citep{DBLP:journals/corr/abs-1911-06394}, improving zero-shot DST performance across all domains to 11\% joint accuracy over the state-of-the-art.
\end{itemize}

\section{Approach}
We follow \citet{DBLP:conf/sigdial/GaoSACH19} in treating Dialog State Tracking as a reading comprehension problem: at each turn of dialog, our model reads the dialog history and answers a fixed set of queries about user requirements and preferences (slots), with predictions aggregated to form the belief state.
In our framework (\Cref{fig:problem_setup}), we query for a given slot (e.g.~Hotel Price Range) by asking a natural language question \citep{DBLP:journals/corr/abs-2004-05827}---``What is the price range of the hotel the user prefers?".
As our model's predictive ability is based on its general understanding of language and task-oriented conversation, we support zero-shot inference without the need to re-train the model or extend a formal ontology.
For example, if a model has not been trained on data from the hotel domain, when presented with a hotel booking conversation we may nonetheless ask it a question like ``In what area is the user looking for a hotel?" and received a prediction for that unseen requirement (Hotel Area).

While we conduct our experiments on English-language DST datasets, our approach is applicable to state tracking in any language, provided a conversation history is available.

\paragraph{Problem Statement}
We consider a conversation with $T$ turns of user $u_t$ and system utterances $y_t$: $C = \{ y_1, u_1, \dots y_T, u_T \}$.
The belief state $B_t$ at turn $t$ comprises many tuples of slots $s \in S$ and their associated values $v_{s,t} \in V_s$, extracted from the conversation history $C_t = \{y_1, u_1, \dots, y_t, u_t \}$.
The set of possible values $V_s$ can be arbitrarily large (e.g.~possible hotel names), so we represent these values as sequences of vocabulary tokens $v_{s,t} = \{w_1, w_2, \dots, w_k \}, w_i \in \mathcal{W}$.
At inference time we pose a natural language question $s = \{ w_1, \dots, w_n \}$ and our model predicts an answer (slot value $v_{s,t}$) based on its understanding of the dialog history $C_t$.
To predict the belief state $B_t$, our model independently answers $|S|$ different questions (\Cref{fig:problem_setup}).
In zero-shot DST, the system must predict values for slots outside of the initial ontology---these slot queries correspond to arbitrary natural language questions $s'$ about entities and relationships in the conversation $C_t$.

\begin{figure}[t!]
\centering
\includegraphics[width=0.99\columnwidth]{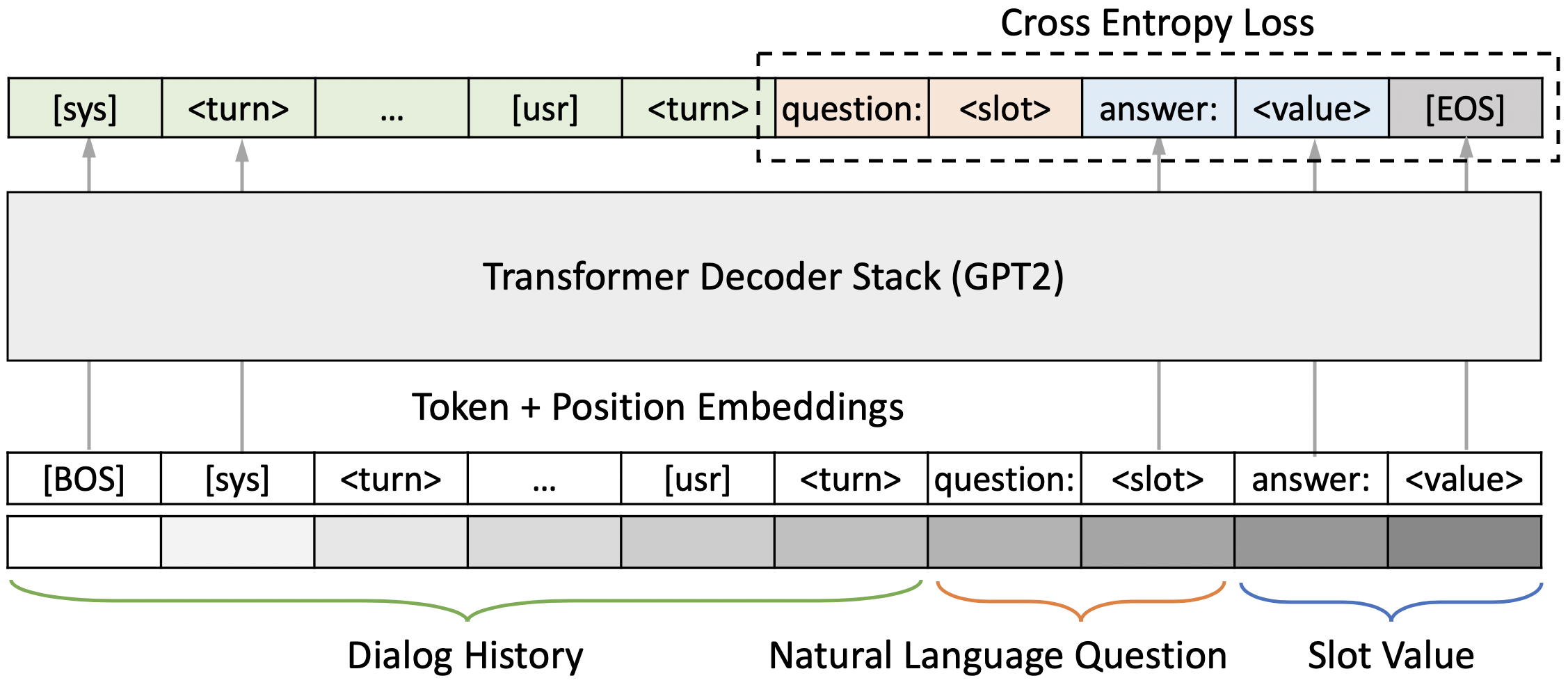}
\caption{Our model performs DST via generative question-answering. Natural language questions for dialog slots allow our model to generalize to new slot types through its understanding of general language.}
\label{fig:model}
\end{figure}

\paragraph{Generalizing to New Domains and Slots}
Dialog State Tracking systems in real-world settings must scale to new users and services, accommodating new slot values (e.g.~a new movie release) as well as new domains and slot types (e.g.~a service update, or a new connected API).
Existing methods require the developer to either write a complete ontology of slots and allowed values or modify their model architecture to add slot-specific prediction heads \citep{DBLP:conf/aaai/0002LWZT020}.
Span-based approaches \citep{DBLP:journals/corr/abs-1910-03544,DBLP:journals/corr/abs-1911-06192} can correctly predict values that appear verbatim in a conversation but fail when a user paraphrases or mis-phrases a value.
They also fall back to treating open-valued slots as classification problems \citep{DBLP:journals/corr/abs-1910-03544,DBLP:journals/corr/abs-2004-05827}.
We approach DST as an ontology-free generative question answering task, as generative methods \citep{DBLP:conf/acl/WuMHXSF19, DBLP:conf/aaai/KumarKGMH20} have shown promise in few-shot and supervised DST settings.

While some approaches toward DST as QA learn a set of embeddings for each slot and/or domain \citep{DBLP:conf/sigdial/GaoSACH19,DBLP:conf/acl/WuMHXSF19, DBLP:conf/aaai/KumarKGMH20}, this is not robust to unseen slots.
We encode slots as natural language questions---manually formulating one question per slot---allowing us to share a pre-trained encoder for both dialog context and slot to leverage shared linguistic knowledge \citep{DBLP:journals/corr/abs-2004-05827}.
Thus, our model is also agnostic to ontologies and can answer arbitrary English questions about the dialog history.
We treat DST via QA as a conditional language modeling task, and train our model to predict the conditional likelihood of question (slot $s$) and answer (value $v_{s, t}$) tokens given a dialog context $C_t$ at a given turn $t$:
\begin{align*}
    P(v_{s,t}, s| C_t) = P(v_{s,t} | s, C_t) * P(s | C_t)
\end{align*}
At inference time, the model is given the dialog context alongside a question---$[C_t; s]$---and asked to predict the value $v_{s,t}$ for that slot.

\begin{table}[t!]
\centering
\small
\begin{tabular}{@{}lrr@{}}
\toprule
                       & JGA (\%)            & \# Params \\ \midrule
DistilGPT2 LM          & 36.35               & 82 M      \\
DistilGPT2 CLM no PT   & 39.34               &       \\
DistilGPT2 CLM         & 49.55               &       \\
\quad +Question (CLMQ)        & \underline{50.83}   &           \\\midrule
GPT2 CLMQ              & 51.02               & 124 M     \\ \midrule
GPT2-medium CLMQ       & \textbf{52.58}      & 355 M     \\ \bottomrule
\end{tabular}
\caption{Ablation study of our framework, reporting supervised JGA on the MultiWOZ 2.1 test set.}
\label{tab:supervised-ablation}
\end{table}

\section{Model Architecture}
For our conditional language model, we compared two common architectures: 
1) an encoder-decoder model \citep{DBLP:conf/nips/SutskeverVL14} with a bi-directional encoder; and 2) a purely auto-regressive decoder-only model.
We conducted preliminary experiments using both a Transformer \citep{DBLP:conf/nips/VaswaniSPUJGKP17} encoder-decoder language model pre-trained using a de-noising auto-encoder objective \citep{DBLP:conf/acl/LewisLGGMLSZ20}, as well as a Transformer decoder pre-trained with next-token prediction on English web pages.
We achieved 1\% better supervised DST performance with the decoder-only model in half the training time.
Our model architecture thus comprises a Transformer decoder language model that allows us to leverage pre-trained language models like GPT2 \citep{radford2019better} and common-sense world knowledge accrued through pre-training \cite{DBLP:conf/emnlp/PetroniRRLBWM19}.

We use a BPE \citep{DBLP:conf/acl/SennrichHB16a} tokenizer to convert input text into a sequence of tokens.
These are embedded in $\mathbb{R}^h$ and added to an $\mathbb{R}^h$ sinusoidal positional embedding.
This input embedding is processed by $l$ Transformer layers with hidden dimensionality $h$, each of which applies multi-headed attention with $k$ heads followed by a feed-forward layer with a softmax nonlinearity.
The final output hidden states are then projected into our vocabulary space of 50,257 sub-word tokens.
We initialize our model weights with DistilGPT2 \citep{DBLP:journals/corr/abs-1910-01108}, GPT2 \citep{radford2019better}, or GPT2-medium with $h=768, 768, 1024$, $l=6, 12, 24$, and $k=12, 12, 16$ respectively.

\begin{table}[t!]
\small
\centering
\begin{tabular}{@{}lrr@{}}
\toprule
                  & MultiWOZ & DSTC8  \\ \midrule
Train             & 7,906    & 16,142 \\
Validation        & 1,000    & 2,482  \\
Test              & 1,000    & 4,201  \\ \midrule
Domains           & 5        & 19     \\ \midrule
Slots             & 30       & 124    \\
\quad Open        & 9        & 59     \\
\quad Numeric     & 5        & 12     \\
\quad Temporal    & 5        & 10     \\
\quad Categorical & 11       & 43     \\ \bottomrule
\end{tabular}
\caption{Dataset statistics for MultiWOZ 2.1 and DSTC8: number of dialogs in each split, number of domains, and slots with slot category breakdowns.}
\label{tab:dataset-stats}
\end{table}

As seen in \Cref{fig:model}, our input sequence consists of a concatenation of dialog context $C_t$, slot query $s$, and slot value $v_{s,t}$: $[C_t; s; v_{s,t}]$.
We pre-pend each utterance with a speaker token \texttt{[usr]} or \texttt{[sys]} for a user or system speaker to allow our model to identify additional context about each utterance.
We pre-pend the slot query and value with \texttt{question:} and \texttt{answer:}  respectively to distinguish slot queries from user-posed questions in the conversation.
At training time, we calculate a cross-entropy loss similar to encoder-decoder models by maximizing the log likelihood of the slot query and value conditioned on the dialog context:
\begin{align*}
    P(s, v_{s,t}|C_t) = \prod^n_i{P(x_i|x_{<i}, C_t)}
\end{align*}
where $n=|[s; v_{s,t}]|$.
We find through ablation experiments on our architecture that this loss computation method out-performs a na\"ive language-modeling approach that maximizes log likelihood of the full concatenated sequence $[C_t; s; v_{s,t}]$ via the factorized joint distribution \citep{DBLP:journals/corr/abs-2005-05298, DBLP:journals/corr/abs-2005-00796}:
\begin{align*}
    P(x) = \prod^n_i{P(x_i|x_{<i})}
\end{align*}
This allows for flexibility in learned representations for dialog context while regularizing slot query hidden states.

\section{Data}

We perform our experiments on \textbf{MultiWOZ} \citep{DBLP:conf/emnlp/BudzianowskiWTC18}, which contains over 10K single- and multi-domain task-oriented dialogs written by crowd-workers.
We use the 2.1 version, with corrected and standardized annotations from \citet{DBLP:conf/lrec/EricGPSAGKGKH20}.
We follow \citet{DBLP:conf/acl/WuMHXSF19}
in lower-casing all dialogs and removing dialogs from training-only domains (Police and Hospital).
The final dataset contains 9,906 conversations from 5 domains (Restaurant, Hotel, Attraction, Train, Taxi) covering 30 domain-slot pairs.
Each dialog contains an average of 7 user and system turns.

\begin{table}[t!]
\small
\centering
\begin{tabular}{@{}lcrc@{}}
\toprule
Model         & Type       & JGA            & NLQ       \\ \midrule
TRADE \citep{DBLP:conf/acl/WuMHXSF19}  & G          & 45.60          &            \\
SUMBT \citep{DBLP:conf/acl/LeeLK19}*        & C          & 46.70          &            \\
STARC \citep{DBLP:journals/corr/abs-2004-05827}*        & C+S        & 49.48          & Y          \\
MA-DST \citep{DBLP:conf/aaai/KumarKGMH20}    & G          & 51.88          &            \\
\underline{GPT2-m CLMQ} & G & \textbf{52.58} & Y \\ \bottomrule
\end{tabular}
\caption{Supervised DST performance on MultiWOZ 2.1 of our model (\underline{underlined}) compared to prior methods capable of zero-shot inference.
Models using natural language questions (NLQ) are marked. *\emph{Requires access to slot-value ontologies at inference time.}}
\label{tab:supervised-zs}
\end{table}

We also experiment with augmenting our training dataset in zero-shot settings with observations drawn from the DSTC8 \citep{DBLP:journals/corr/abs-1911-06394} dataset,
\footnote{\url{https://github.com/google-research-datasets/dstc8-schema-guided-dialogue}}
which contains 16,152 dialogs from 45 domains.
DSTC8 was created via template-based dialog models provided with service APIs, and then edited by crowd-workers \citep{DBLP:journals/corr/abs-1801-04871}.
We normalize domains and slots corresponding to the same domain (e.g.~\texttt{Bus\_1, Bus\_2}) for a total of 19 domains and 124 slot types in DSTC8.
We further manually annotate each dataset with slot value types: open-valued (e.g.~Hotel Name), numeric (e.g.~Restaurant Guests), temporal (e.g.~Taxi LeaveAt), and categorical (e.g.~Attraction Type).
Dataset statistics are shown in \Cref{tab:dataset-stats}.

\section{Experiments}

We measure DST performance via Joint Goal Accuracy (JGA): the proportion of turns with all belief slots predicted correctly, including those not present.
In \Cref{supervised}, we evaluate our model on fully \emph{supervised} DST, in which all domains and slots are known at training time.
In \Cref{zs}, we investigate \emph{zero-shot} domain adaptation in which the model is evaluated on conversations from an unseen domain with previously unseen slots.
We then explore how our framework seamlessly accommodates teaching a model to predict slot carry-over (\Cref{carryover}) and transfer learning with significantly more diverse domains and slot types (\Cref{transfer}).
To measure zero-shot JGA, we follow \citet{DBLP:conf/acl/CampagnaFML20} and only consider slots specific to the held-out domain.
We focus our analysis on the zero-shot setting, as our goal is to build DST systems that can easily and effectively generalize to new domains and services.
We train all models to convergence with a maximum of 10 epochs on Nvidia V100 GPUs, using the Lamb optimizer \citep{DBLP:conf/iclr/YouLRHKBSDKH20} with a base learning rate of 2e-5.
All predictions are made using greedy decoding.

\begin{table}[t!]
\small
\centering
\begin{tabular}{@{}lcrl@{}}
\toprule
Model         & Type       & JGA            & Extra Supervision       \\ \midrule
DSTQA         & C+S        & 51.17          & Knowledge Graph         \\
DS-DST        & C+S        & 51.21          &                         \\
\underline{GPT2-m CLMQ} & G & 52.58 &               \\
SOM-DST       & G          & 53.68          & Previous Dialog State   \\
SST           & C          & 55.23          & Schema                  \\
TripPy        & S          & 55.30          & Previous Dialog Actions \\
SimpleToD     & G          & \textbf{55.72}          &    Actions (Training)                     \\ \bottomrule
\end{tabular}
\caption{Supervised DST performance on MultiWOZ 2.1 of our model (\underline{underlined}) against state-of-the-art DST methods incapable of zero-shot inference.
}
\label{tab:supervised-non-zs}
\end{table}

\subsection{Supervised DST}
\label{supervised}

We first evaluate on the commonly benchmarked supervised DST task to demonstrate performance competitive with state-of-the-art.
In this setting we compare our approach against prior methods capable of zero-shot inference in \Cref{tab:supervised-zs}---TRADE, STARC, SUMBT, and MA-DST---and those incapable of doing so in \Cref{tab:supervised-non-zs}, including DSTQA \citep{DBLP:journals/corr/abs-1911-06192}, DS-DST \citep{DBLP:journals/corr/abs-1910-03544}, SOM-DST \citep{DBLP:conf/acl/KimYKL20}, SST \citep{DBLP:conf/aaai/0002LWZT020}, TripPy \citep{DBLP:journals/corr/abs-2005-02877}, and SimpleToD \citep{DBLP:journals/corr/abs-2005-00796}.
Our model outperforms all prior models that support zero-shot generalization and is competitive with methods that focus solely on supervised DST---most of which require extra supervision at training and inference time, including dialog actions and prior dialog states.
We distinguish models by their prediction type as (C)lassification-, (S)pan extraction-, and (G)eneration-based methods.

As seen in \Cref{tab:supervised-ablation}, our formulation of DST as a generative QA task benefits significantly from the usage of a conditional decoder-style model.
A standard auto-regressive language modeling formulation (\textbf{LM}) with loss computed over the entire input sequence achieves 13\% lower JGA compared to computing cross entropy loss only over slot value tokens (\textbf{CLM}).
Pre-training is also crucial---we see a 10-point drop in JGA when randomly initializing model weights (\textbf{no PT}) compared to initializing from pre-trained DistilGPT2 weights.
We also compare two other sizes of our models: GPT2-based---comparable in size to SUMBT's \citep{DBLP:conf/acl/LeeLK19} 112M parameters---and GPT2-medium-based---comparable in size to STARC's \citep{DBLP:journals/corr/abs-2004-05827} 355M parameters.
We find that scaling the size of our model results in modest improvements in supervised JGA.
We hypothesize that extending our loss to cover both slot query and value tokens (\textbf{+Question/CLMQ}) helps regularize the hidden representations of question tokens, and we achieve a 1.3\% improvement in JGA.

\begin{table}[t!]
\small
\centering
\begin{tabular}{@{}lrrrrr@{}}
\toprule
            & Rest.          & Hot.           & Attr.          & Train          & Taxi                 \\ \midrule
TRADE       & 12.59          & 14.20          & 20.06          & 22.39          & 59.21 \\
MA-DST      & 13.56          & 16.28          & 22.46    & \underline{22.76}    & 59.27 \\
SUMBT       & \underline{16.50}    & \underline{19.80}    & \underline{22.60}          & 22.50       & \underline{59.50}                \\\midrule

Ours (GPT2) & 21.05          & 18.54          & 23.67          & 24.34          & 59.10                \\
Ours (GPT2-m) & \textbf{26.17} & \textbf{24.41} & \textbf{31.31} & \textbf{29.07} & \textbf{59.61}                \\ \bottomrule
\end{tabular}
\caption{Zero-shot domain adaptation JGA (\%) on MultiWOZ 2.1 test set for recent works and our models with question loss, on the (Rest)aurant, (Hot)el, (Attr)action, Train, and Taxi domains. Previous state-of-the-art results are \underline{underlined}, with new best \textbf{bolded}.}
\label{tab:zs-size}
\end{table}

\subsection{Zero-Shot DST}
\label{zs}

Our primary focus lies in the zero-shot domain adaptation setting, where conversations and target slots at inference time come from unseen domains.
We use a leave-one-out setup, training our models on four domains from MultiWOZ and evaluating on the held-out domain.
Our model must understand a wide variety of possible questions about unseen conversations to generalize well.
We compare our model against strong baseline models for zero-shot DST: TRADE, SUMBT, and MA-DST; \Cref{tab:zs-size} contains results from our models alongside baseline results reported by \citet{DBLP:conf/aaai/KumarKGMH20} and \citet{DBLP:conf/acl/CampagnaFML20}.
These models represent slots as domain-slot tuples: TRADE learns a separate embedding for each domain and word in slot names, while SUMBT and MA-DST encode domain-slot tuples via BERT \citep{DBLP:conf/naacl/DevlinCLT19} and an RNN encoder, respectively.

Our GPT2-medium based model achieves state-of-the-art zero-shot performance on all five domains, and by a significant (5-10\%) margin on the Restaurant, Hotel, Attraction, and Train domains.
While increased model size modestly impacts supervised DST performance (\Cref{tab:supervised-ablation}), larger models perform significantly better in a zero-shot setting with average absolute gains of 4.8\% and relative gains of 22\% in JGA across domains.
Such improvements are consistent with findings from \citet{DBLP:journals/corr/abs-2005-14165} that up-sizing language models improves zero-shot performance across various tasks and \citet{DBLP:conf/emnlp/PetroniRRLBWM19}, who observe that larger pre-trained models can retain more common-sense and world knowledge from their pre-training corpus---which may help our model understand queries for unseen domains and slots.

\paragraph{Effect of Natural Language Questions}

Prior work that frames DST as QA typically represents the slot query as a concatenation (tuple) of domain and slot name.
\citet{DBLP:journals/corr/abs-1910-03544} explore the impact of three different slot representations---domain-slot tuples, short slot descriptions, and full questions---on a hybrid classification-extraction model for DST, and find little difference in performance.
However, we find that full questions work much better than domain-slot tuples for our generative framework, especially in zero-shot DST.
We hypothesize that natural language questions---structurally similar to dialog utterances and pre-training sentences---allow our model to best leverage its linguistic knowledge with minimal friction when jointly encoding the dialog history, slot query, and slot value.

\begin{table*}[t!]
\centering
\small
\begin{tabular}{@{}llllrrrr@{}}
\toprule
USER         & \multicolumn{7}{l}{My friend told me about Carolina Bed and Breakfast. Do you know anything about it?} \\
SYS           & \multicolumn{7}{l}{It's a 4 star guesthouse. What would you like to know about it?}      \\ 
USER          & \multicolumn{7}{l}{Can you give me the postcode? And, do they have internet?}            \\
SYS           & \multicolumn{7}{l}{The postcode is cb13nx; they have internet.}                          \\
USER          & \multicolumn{7}{l}{Thanks. Any boat attractions in the west?}                            \\
\textbf{SYS} & \multicolumn{7}{l}{\textbf{Nothing in west. Closest boat is the Cambridge Punter in centre. Too far?}} \\ 
\textbf{USER} & \multicolumn{7}{l}{\textbf{Yes, it is. How about a museum?}}                             \\ \midrule
Error Modality        & Slot            & Gold   & Prediction       & Open    & Numeric & Temporal & Categorical \\ \midrule
Spurious      & (Attraction, Name) & \texttt{n/a}    & cambridge punter & 8.4 \%  & 22.3 \% & 47.7 \%  & 16.0 \%     \\
Ignored       & (Hotel, Internet)  & yes    & \texttt{n/a}              & 65.3 \% & 53.5 \% & 19.9 \%  & 76.8 \%     \\
Wrong Value    & (Attraction, Type) & museum & boat             & 26.3 \% & 24.2 \% & 32.4 \%  & 7.2 \%      \\ \bottomrule
\end{tabular}
\caption{Example of different classes of DST errors, and the proportion of errors they make up across the four slot categories for all five domains in a zero-shot setting. Latest (target) turn is \textbf{bolded}.}
\label{tab:error-ex}
\end{table*}

\citet{DBLP:conf/acl/WuMHXSF19} find that zero-shot generalization in models that represent slots as tuples is primarily due to shared slot names between domains (e.g.~Taxi and Train `leaveAt').
In a real-world setting a newly added dialog service is unlikely to share slot names verbatim with existing services.
To fairly compare tuples and natural language questions under our framework, we perform zero-shot experiments using each representation.
For tuple-based questions, our model takes as slot query a synonym of the slot name (e.g.~Taxi `leaveAt' $\rightarrow$ `Pick Up Time') instead of a full question (e.g.~`What time does the user want the taxi to pick them up?`).
Full question models achieved 6\% higher per-domain JGA compared to slot-tuple models, supporting the notion that slot-tuple models memorize slot names rather than understanding their meaning and thus do not generalize well in real-world settings.
Using full questions, our model (\Cref{tab:zs-size}) achieves state-of-the-art performance in zero-shot settings.

\paragraph{Error Modalities}

To analyze our model, we follow \citet{DBLP:journals/corr/abs-2004-05827} and 
categorize DST errors in three modalities: 1) the model predicts a \emph{spurious} value for an irrelevant slot; 2) the model \emph{ignores}
a relevant slot; and 3) the model correctly infers the presence of a slot but predicts a \emph{wrong value}.
\Cref{tab:error-ex} shows examples of each type of error for a sample conversation, and what proportion of errors they make up in each slot category for our \textbf{GPT2-m CLMQ} model in a zero-shot setting.
Temporal slots are least likely to be \emph{ignored} by our model, as verbatim \texttt{HH:MM} values are easily identifiable in a conversation.
However, it is difficult to distinguish between closely related unseen temporal slots like `leaveAt' and `arriveBy'.
Values for categorical, numeric, and open-valued slots on the other hand can comprise common (non-slot) phrases used in conversation, and thus it is easy for our model to ignore such slot references.

We also examine the source of dialog slots: \emph{users} explicitly express the majority (79.5\%) of slot values, while a minority are either derived via user reactions to \emph{system} suggestions (9.7\%) or \emph{implicitly} valued (10.8\%)---not present verbatim in a conversation.
However, our errors are distributed evenly between \emph{user}, \emph{system}, and \emph{implicit} sourced slots---suggesting that it is challenging for our model to track dialog states that are updated reactively via user feedback.
We thus see a future opportunity to improve DST models by emphasizing multi-hop reasoning and common-sense inference.

\subsection{Predicting Carried Over Slots}
\label{carryover}

Long-range dependencies and slot values carried over from early turns are particularly important to model for accurate DST in long conversations \cite{DBLP:conf/aaai/KumarKGMH20}.
We observe this in the zero-shot setting: our model is able to predict all slots accurately for 61\% of conversation first-turns, dropping to 46\% after one turn, and 5.7\% after seven turns (the average conversation duration).
We implement an oracle module to discard predictions when a dialog state does not need updating, obtaining an upper bound for DST improvements due to carry-over prediction.
With this oracle, we see an average 5-point improvement in JGA across domain,
indicating that carry-over prediction can greatly benefit our model.
State-of-the-art models for fully supervised DST 
often rely on explicitly processing previous dialog states---via slot-value graphs \citep{DBLP:journals/corr/abs-1911-06192, DBLP:conf/aaai/0002LWZT020} or as a separate input to the model at each turn \citep{DBLP:journals/corr/abs-2005-02877, DBLP:conf/acl/KimYKL20}.
In our framework we can target slot carry-over by training a model to predict a \texttt{carried over} token in place of the true slot value whenever a slot value does not need updating at the current turn (\textbf{+ Carryover}).
At inference time, we replace predicted carry-over tokens with the slot's last predicted value.

\begin{figure}[t!]
\centering
\includegraphics[width=0.99\columnwidth]{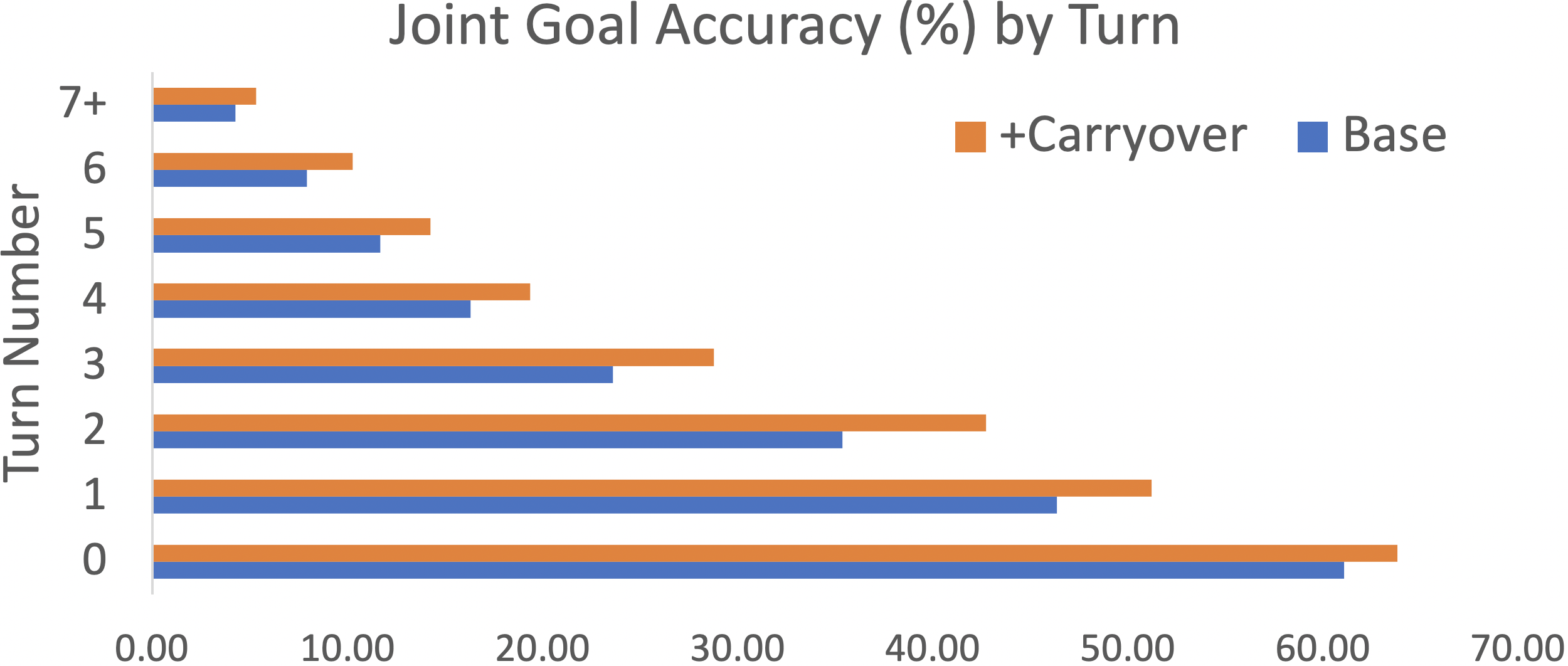}
\caption{Per-turn JGA on MultiWOZ Test set for GPT2-CLMQ with and without carry-over prediction.}
\label{fig:turn_jga}
\end{figure}

Our carry-over implementation improved JGA for all domains (\Cref{tab:perf-modifications}) by an average of 3.14\%, and improved JGA across all context lengths---with the largest improvements (+7\%) at the second and third turn of a conversation (\Cref{fig:turn_jga}).
The \texttt{carried over} token allows our model to hedge against low confidence slots, falling back to predictions from previous turns where the target slot may be directly mentioned.
This helps reduce the \emph{wrong value} error rate by an average of 31\% across each domain.
Our model can also propagate null values with carry-over, reducing spurious predictions by an average of 36\% across domains.
However, we also observe our carry-over model propagating 78\% of its errors from previous turns, suggesting that further improvements can result via accurately predicting slot updates.

\begin{table}[t!]
\centering
\small
\begin{tabular}{@{}lrrrrr@{}}
\toprule
              & Rest.          & Hotel          & Attr.          & Train          & Taxi           \\ \midrule
Previous SOTA & 16.50          & 19.80          & 22.60          & 22.76          & 59.50          \\ \midrule
GPT2 CLMQ     & 21.05          & 18.54          & 23.67          & 24.34          & 59.10          \\
\quad + Carryover & 24.00          & 19.91          & 28.45          & 30.75          & 59.29 \\
\quad + DSTC8      & \underline{24.65} & \underline{22.94} & \underline{34.30} & \underline{38.55} & \underline{59.68}          \\ \midrule
GPT2-m CLMQ      & 26.17          & 24.41          & 31.31          & 29.07         & 59.68          \\
\quad + CO, DSTC8      & \textbf{27.69}          & \textbf{24.88}          & \textbf{42.39}          & \textbf{41.05}         & \textbf{60.32}          \\ \bottomrule
\end{tabular}
\caption{Zero-shot JGA on MultiWOZ 2.1 test set with carry-over prediction and transfer learning.}
\label{tab:perf-modifications}
\end{table}

\subsection{Transfer Learning for Generalization}
\label{transfer}

Our framework is ontology-agnostic and thus easily supports transfer learning without modifying the architecture by simply writing natural language questions for additional slots.
\citet{DBLP:journals/corr/abs-2004-05827} found that intermediate fine-tuning of RoBERTa-Large \citep{DBLP:journals/corr/abs-1907-11692} on passage-based QA tasks \citep{DBLP:conf/acl-mrqa/FischTJSCC19} improved zero-shot DST performance.
In preliminary experiments, we found no significant impact from intermediate fine-tuning on the SQuAD v2.0 \citep{DBLP:conf/acl/RajpurkarJL18} passage-based QA dataset.
However, we observe significant improvements when training with joint, non-curriculum learning \citep{DBLP:journals/corr/abs-1806-08730, DBLP:journals/corr/abs-1910-10683}---augmenting our training data with an equal number of examples sampled from DSTC8, taking care to remove data from the held-out domain in both MultiWOZ and DSTC8.

Our framework allows for easy joint optimization with carry-over and transfer learning: by training new models on MultiWOZ 2.1 augmented with DSTC8 (\textbf{+ DSTC8}) we gain a further average 3.5-point improvement in per-domain JGA (\Cref{tab:perf-modifications}).
On average, our model makes 29\% fewer \emph{spurious} errors, and 6.9\% fewer errors in open-valued slots,
suggesting that our model scales well with additional training data with semantically distinct slot types and values.
Our model also makes 9.7\% fewer errors on categorical slots and 63\% fewer mistakes where it assigns the value of one categorical slot to another, despite being unable to observe the set of possible categorical options---suggesting that exposure to more diverse categorical slots allows our model to better understand and distinguish between such slots.
While temporal slots comprise only 17\% of MultiWOZ and 10\% of DSTC8 slots, these additional examples seem to help our model better disambiguate temporal references and make 32\% fewer errors in such slots.

By applying both carry-over and transfer learning to our largest model, we observe further improvements in zero-shot JGA for all domains---averaging 5.1 points better than GPT2-m CLMQ, for an average gain of 11\% JGA over previous state-of-the-art across domains (\Cref{tab:perf-modifications}).

\begin{figure}[t!]
\centering
\includegraphics[width=0.99\columnwidth]{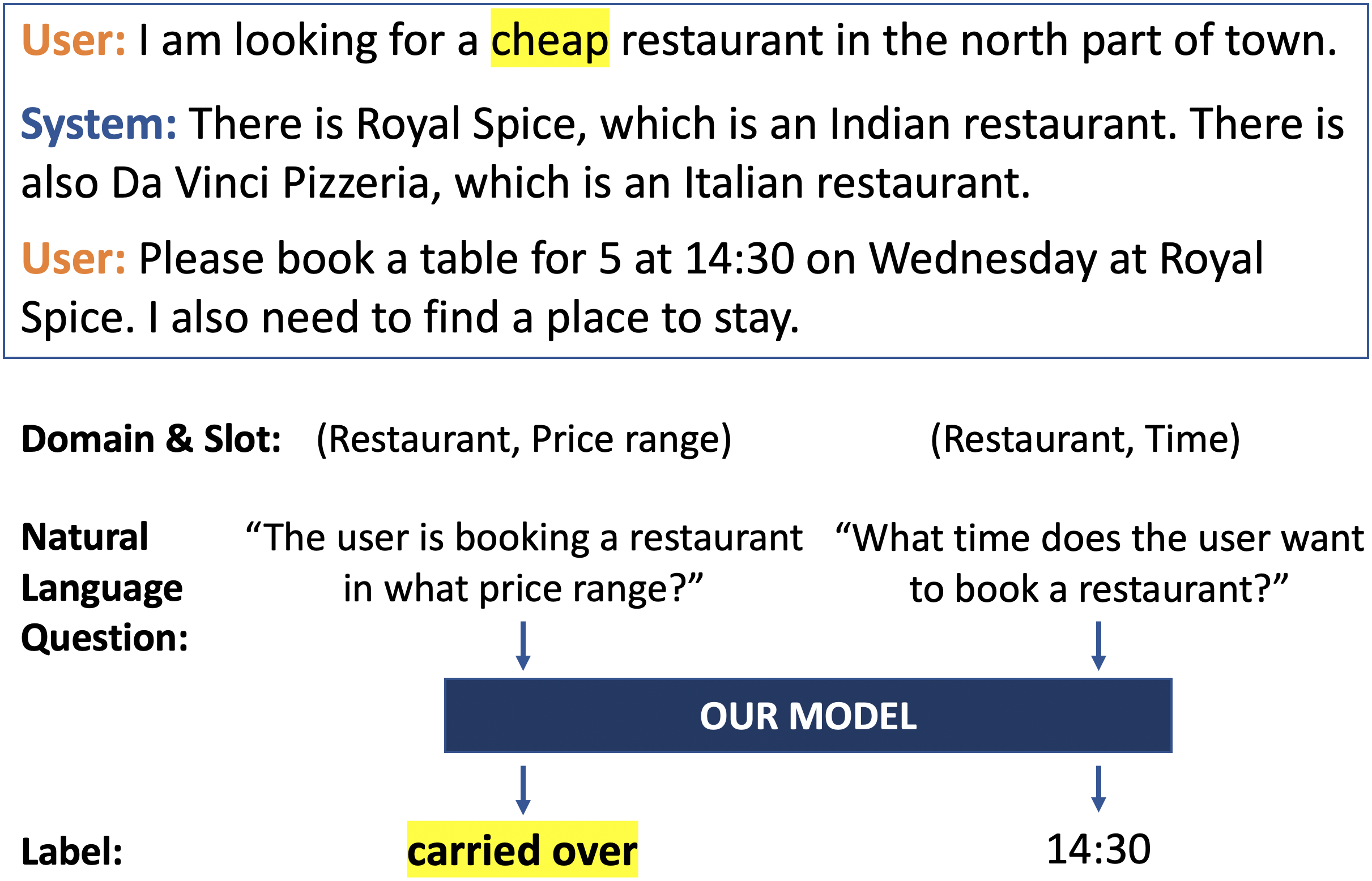}
\caption{Our model can be trained to predict the presence of slot carry-over by replacing slot values from previous turns with the \texttt{carried over} token.}
\label{fig:carryover_setup}
\end{figure}

\section{Qualitative Analysis}

We manually reviewed 300 errors made by our GPT2-medium CLMQ model in the zero-shot setting---annotating 20 errors from each modality (spurious, ignored, wrong value) from each domain with the gold label quality and perceived cause of error totaling 300 annotated examples.
As widely observed in recent DST work \citep{DBLP:journals/corr/abs-1911-06192, DBLP:conf/aaai/KumarKGMH20}, a significant proportion of DST errors on MultiWOZ are unavoidable---caused by annotation errors.
While version 2.1 corrected some of these, annotation errors and inconsistencies remain responsible for 30\% of sampled errors---in particular, in 10\% of errors the original annotator did not record reactive preferences while in 5\% of errors the original annotator did.
These inconsistencies can hurt our model's ability to infer reactive and implied requirements and preferences.

We are also particularly interested in \emph{slot transfers}---when our model mistakenly predicts one slot's value for a different slot, comprising 36\% of our manually reviewed errors.
In the Taxi and Hotel domains, our model transfers slots from the same domain over 75\% of the time, with most swaps occurs between same-category slots (e.g.~temporal slots like Taxi `LeaveAt' and `ArriveBy').
Slots in these domains are closely semantically related, with values that can fit any slot of that category (e.g.~13:10 vs. 15:15).
While a human can easily infer that the earlier of two times must be departure and the later arrival, our model has no inherent understanding of temporal mechanics or numeracy \cite{DBLP:conf/emnlp/WallaceWLSG19}.
In future work, we will explore learning such knowledge directly via hierarchical softmax output distributions to distinguish between output modalities \cite{DBLP:conf/acl/RiedelS18}, and fine-tuning our model with contrastive losses to learn to rank numerals and times \cite{DBLP:journals/corr/HofferA14}.

For Restaurant, Attraction, and Train, our model tends to swap slot values with those from other domains in the conversation.
This is often due to semantically similar slots whose values, at first glance, may not be obviously identifiable as such (e.g.~`Bridge' or `The Place').
\citet{DBLP:conf/aaai/KumarKGMH20} similarly observe a particularly high incidence of slot transfers between different-domain `Name' slots.
Other such slots include price ranges and numbers of guests.
We have seen that data augmentation with DSTC8 can improve our model's ability to disambiguate such slots---this suggests that we could further improve our model by exposing it to in-domain, conversational reading comprehension data.

While no such dataset currently exists, in future work we aim to explore using question generation \cite{DBLP:conf/acl/DuSC17} and paraphrasing \cite{DBLP:conf/taai/TsengHCS14} models to perform in-domain data augmentation, creating reading comprehension questions for task-oriented dialogs that targeting entities and relations not covered by an ontology.
We also wish to explore methods for generating general reading comprehension questions for out-of-domain conversations \cite{DBLP:conf/emnlp/ShakeriSZNNWNX20} to improve our model's domain adaptation ability.

\section{Related Works}

Modern dialog state tracking seeks to capture evolving user intents in a structured belief state \citep{DBLP:journals/csl/ThomsonY10}.
Traditional systems rely on hand-crafted features \citep{DBLP:conf/sigdial/HendersonTY14} and classify slot values from a fixed ontology \citep{DBLP:conf/acl/MrksicSWTY17,DBLP:conf/acl/RamadanBG18}.
\citet{DBLP:conf/sigdial/GaoSACH19} and \citet{DBLP:journals/corr/abs-1911-06192} fill some slots via spans extracted from dialog history, although they treat non-numeric slots as categorical.
Generative methods \citep{DBLP:conf/acl/XuH18,DBLP:conf/acl/WuMHXSF19} can predict arbitrary unseen values, with \citet{DBLP:journals/corr/abs-2005-00796} achieving state-of-the-art supervised DST performance in MultiWOZ 2.1 although they cannot predict unseen slots.

By posing DST as generative QA, our framework can leverage language models pre-trained on open-domain documents \citep{radford2019better} to understand unfamiliar queries.
Like \citet{DBLP:journals/corr/abs-2004-05827}, we seek to answer natural language questions about each slot.
We contrast our approach to zero-shot DST---which never has access to slots or dialog from the target domain---and that of \citet{DBLP:conf/acl/CampagnaFML20}, who expose their `zero-shot' models to synthetic in-domain conversations that require access to the full ontology of the `held-out' evaluation domain. 

We take inspiration from previous work that frames a wide selection of natural language understanding (NLU) tasks \citep{DBLP:conf/iclr/WangSMHLB19} as QA \citep{DBLP:journals/corr/abs-1806-08730} and span extraction \citep{DBLP:journals/corr/abs-1904-09286}.
While question-answering can be posed as a span extraction task \citep{DBLP:conf/acl/WangL016}, generative approaches have proven successful in answering questions about complex passages \citep{DBLP:conf/acl/FanJPGWA19}.
We use a language modeling approach, taking cues from \citet{DBLP:journals/corr/abs-1910-10683} who demonstrate that a large language model trained on next-token prediction can learn to solve many different NLU tasks posed as text.
Recent work has also shown that large pre-trained language models can generalize to new NLU tasks with few or no examples \citep{DBLP:journals/corr/abs-2005-14165}, and we leverage this alongside world knowledge acquired during the pre-training process \citep{DBLP:conf/emnlp/PetroniRRLBWM19} to build a DST model that is robust to new domains and slot-value ontologies.

\section{Conclusion}

This paper proposes a conditional language modeling approach to multi-domain DST posed as a generative question answering task.
By leveraging natural language questions as state queries, our model can generalize to unseen domains, slots, and values via its understanding of language.
Our model achieves state-of-the-art zero-shot results on the MultiWOZ 2.1 dataset with average per-domain absolute improvements of 5.9\% joint accuracy.
We also demonstrate that our framework is easily extensible to support transfer learning and learning slot carry-over.
In the future, it is worth exploring mechanisms for our model to better understand relative temporal values and general reading comprehension questions from conversations in order to disambiguate semantically similar dialog slots.

\section*{Acknowledgments}
We thank Ben Liu, Maryam Fazel-Zarandi, Anuj Goyal, and anonymous reviewers for providing valuable feedback on this work.
Work was performed during first author's internship at Amazon Alexa AI.
Findings and observations are of the authors only, and do not necessarily reflect the views of Amazon or UCSD.

\bibliography{eacl2021}
\bibliographystyle{acl_natbib}

\appendix

\end{document}